\documentclass{article}
\usepackage{ijcai25}

\usepackage{amsmath,amssymb,amsfonts}
\usepackage{algorithmic}
\usepackage{algorithm}
\usepackage[linesnumbered, ruled, vlined, algo2e]{algorithm2e}
\usepackage{textcomp}
\usepackage{xcolor}
\usepackage{amsfonts}
\usepackage{float}
\usepackage{subfig}
\usepackage{tcolorbox}
\usepackage[T1]{fontenc}
\usepackage{times}
\usepackage{soul}
\usepackage{url}
\usepackage[hidelinks]{hyperref}
\usepackage[utf8]{inputenc}
\usepackage{caption}
\usepackage{graphicx}
\usepackage{amsmath}
\usepackage{amsthm}
\usepackage{booktabs}
\usepackage[switch]{lineno}

\graphicspath{{./figures}}

\newcommand{\ourm}{\text{MT-Core}}
\definecolor{backred}{RGB}{255, 242, 242}
\definecolor{titlered}{RGB}{255, 191, 191}
\definecolor{framered}{RGB}{191, 128, 128}
\definecolor{backgreen}{RGB}{242, 249, 249}
\definecolor{titlegreen}{RGB}{191, 223, 223}
\definecolor{framegreen}{RGB}{128, 159, 159}
\definecolor{backyellow}{RGB}{249, 247, 232}
\definecolor{frameyellow}{RGB}{191, 176, 128}
\definecolor{titleyellow}{RGB}{255, 239, 191}

\def\BibTeX{{\rm B\kern-.05em{\sc i\kern-.025em b}\kern-.08em
    T\kern-.1667em\lower.7ex\hbox{E}\kern-.125emX}}

\title{Multi-granularity Knowledge Transfer for Continual Reinforcement Learning}

\author{
Chaofan Pan$^1$
\and
Lingfei Ren$^1$
\and
Yihui Feng$^{1}$
\and
Linbo Xiong$^1$
\and\\
Wei Wei$^2$
\and
Yonghao Li $^1$
\And
Xin Yang$^1$\\
\affiliations
$^1$Southwestern University of Finance and Economics\\
$^2$Shanxi University\\
\emails
pan.chaofan@foxmail.com,
renlf@swufe.edu.cn,
yihuifeng@foxmail.com,
224081200021@smail.swufe.edu.cn,
weiwei@sxu.edu.cn,
liyonghao@swufe.edu.cn,
yangxin@swufe.edu.cn
}

\begin{document}

\maketitle

\begin{abstract}
    Continual reinforcement learning (CRL) empowers RL agents with the ability to learn a sequence of tasks, accumulating knowledge learned in the past and using the knowledge for problem-solving or future task learning.
    However, existing methods often focus on transferring fine-grained knowledge across similar tasks, which neglects the multi-granularity structure of human cognitive control, resulting in insufficient knowledge transfer across diverse tasks.
    To enhance coarse-grained knowledge transfer, we propose a novel framework called \ourm{} (as shorthand for \textbf{M}ulti-granularity knowledge \textbf{T}ransfer for \textbf{Co}ntinual \textbf{re}inforcement learning).
    \ourm{} has a key characteristic of multi-granularity policy learning: 1) a coarse-grained policy formulation for utilizing the powerful reasoning ability of the large language model (LLM) to set goals, and 2) a fine-grained policy learning through RL which is oriented by the goals.
    We also construct a new policy library (knowledge base) to store policies that can be retrieved for multi-granularity knowledge transfer.
    Experimental results demonstrate the superiority of the proposed \ourm{} in handling diverse CRL tasks versus popular baselines.
\end{abstract}

\section{Introduction}
Reinforcement learning (RL) is a powerful paradigm in artificial intelligence (AI) that enables agents to learn optimal behaviors through interactions with environments.
The capacity to continuously adapt and learn from new tasks without starting from scratch is a defining characteristic of human learning \cite{kudithipudi2022biological} and a desirable trait for RL agents deployed in the dynamic and unpredictable real world.
To achieve this, the research of continual reinforcement learning (CRL, a.k.a. lifelong reinforcement learning) has emerged \cite{rolnick2019experience,kessler2022samestatedifferenttaskcontinual}.
It extends traditional RL by empowering agents with the ability to learn from a sequence of tasks, preserving knowledge from previous tasks, and using this knowledge to enhance learning efficiency and performance on future tasks.
CRL aims to emulate the human capacity for lifelong learning and represents an ambitious field of research that tackles the challenges of long-term, real-world applications by addressing problems of diversity and non-stationarity \cite{khetarpal2022towards}.

\begin{figure}
\begin{small}
        \begin{center}
            \includegraphics[width=0.45\textwidth]{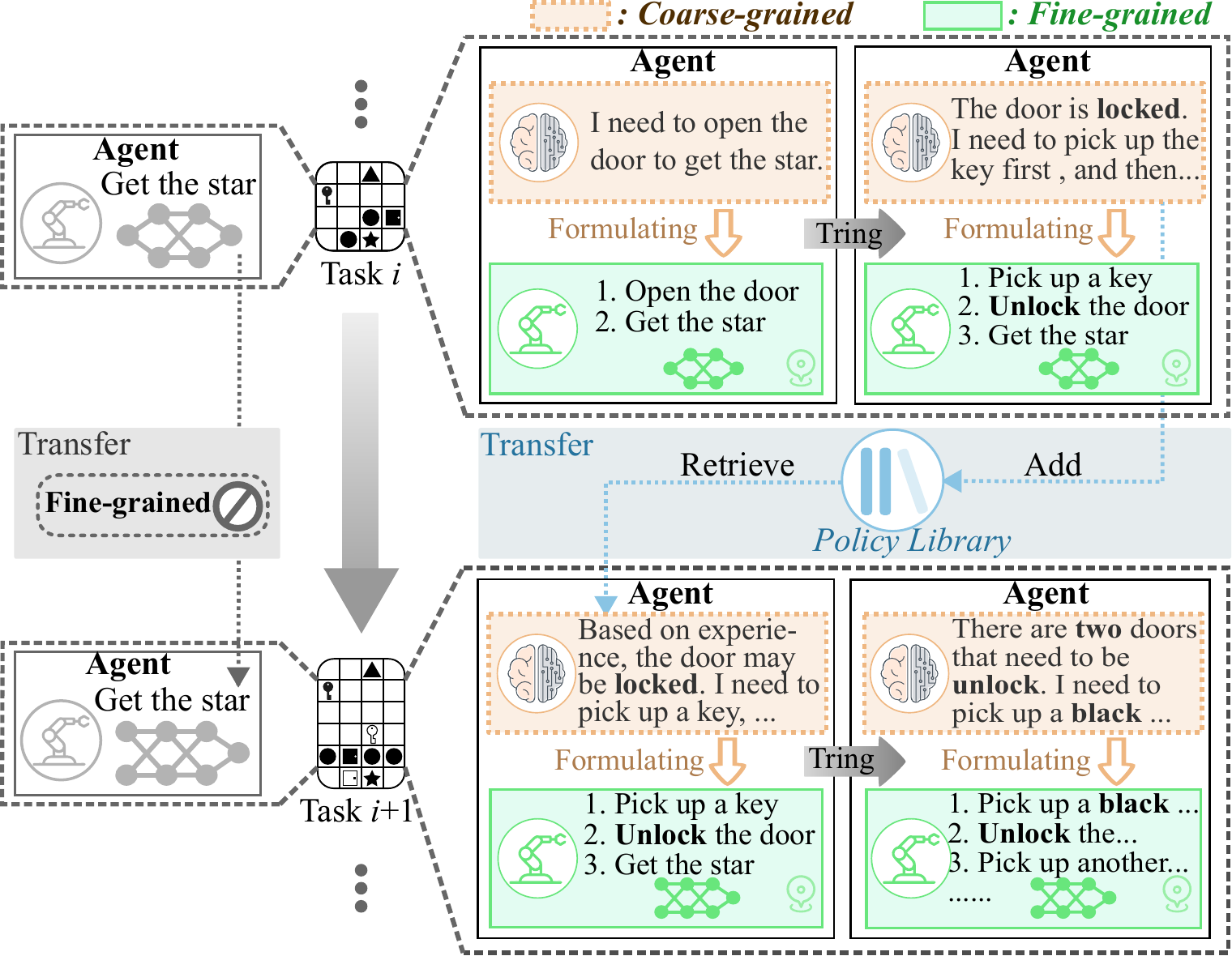}
        \end{center}
        \caption{A simple illustration of our idea.
            The text has been simplified and modified for a better understanding.
            Traditional RL agents (\textbf{left}) primarily focus on fine-grained knowledge transfer, such as sharing the policy network, which may be ineffective across diverse tasks, particularly when the tasks have disparate state spaces.
            In contrast, our agent (\textbf{right}) leverages coarse-grained knowledge transfer within a multi-granularity structure to improve adaptability.
        }
        \label{fig: idea}
    \end{small}
\end{figure}

Despite progress in CRL, the field continues to face a critical challenge: insufficient transfer of knowledge across diverse tasks.
Sufficient knowledge transfer is essential for a robust CRL system, yet many existing methods struggle to achieve this, leading to suboptimal performance and a failure to fully exploit the benefits of continual learning (CL) \cite{wang2023comprehensive,li2024learning}.
A primary reason for this shortfall is the neglect of the sophisticated multi-granularity (hierarchical) structure of human cognitive control, which is primarily located in the prefrontal cortex and is complemented by the cerebellum for precise action control, enabling coarse-grained cognition for envisioning the future and planning \cite{anila2020evidence,friedman2022role}.
This structure enhances humans' ability to tackle complex tasks, and crucially, transfer skills and coarse-grained knowledge across significantly diverse tasks.
In contrast, existing CRL methods often focus on transferring fine-grained knowledge, such as sharing the policy network (left side of Figure \ref{fig: idea}).
This limits them to sequences of highly similar tasks with slight variations only in goals or parameters \cite{kessler2022samestatedifferenttaskcontinual,gaya2023buildingsubspacepoliciesscalablecontinual}, and results in ineffective transfer. 

To take advantage of abstract similarities between diverse tasks, it is essential to represent coarse-grained knowledge effectively and then employ robust models capable of extracting and transferring this knowledge.
Human language, with its inherent abstraction, is a natural fit for representing coarse-grained knowledge.
Furthermore, recent breakthroughs in large-scale language research have demonstrated the impressive reasoning and in-context learning abilities of the large language model (LLM) \cite{zhao2023survey,xi2023rise,wang2023survey}.
Based on these insights, we propose a new framework named \ourm{} (\textbf{M}ulti-granularity knowledge \textbf{T}ransfer for \textbf{Co}ntinual \textbf{re}inforcement learning), which integrates the powerful capabilities of an LLM into the CRL paradigm to enhance coarse-grained knowledge transfer and eventually improve learning performance.

The right side of Figure \ref{fig: idea} illustrates the idea of \ourm{}.
\ourm{} is structured in two layers: the coarse-grained policy formulation and the fine-grained policy learning.
At the coarse-grained, \ourm{} utilizes the powerful reasoning ability of a LLM to formulate a coarse-grained policy, which is represented as a temporally extended sequence of goals.
Each goal consists of a descriptive component and an intrinsic reward function that outlines the intermediate states the agent needs to reach.
The coarse-grained policy is evaluated based on feedback from the fine-grained learning process, allowing the LLM to refine its policy.
Following this process, the coarse-grained policy along with the corresponding information is stored in a policy library.
Upon encountering a new task, \ourm{} can retrieve this policy from the library as a context of the LLM for enhancing coarse-grained knowledge transfer.
At the fine-grained, \ourm{} employs goal-oriented RL to learn a policy that is guided by the formulated coarse-grained policy.
The verified policy can be stored in the policy library to reduce the risk of catastrophic forgetting when encountering old tasks.
Specifically, the coarse-grained policy is represented as \textit{text}, while the fine-grained policy is represented as the policy \textit{network}.
The experimental results demonstrate that \ourm{} significantly outperforms popular baselines.
The main contributions of the work are as follows:
\begin{itemize}
    \item We investigate the knowledge transfer problem in CRL and find that the multi-granularity structure of human cognitive control is crucial for effective knowledge transfer across diverse tasks.
    \item We proposed \ourm{}, a novel CRL framework that leverages multi-granularity knowledge transfer, which is the first work to integrate the powerful reasoning ability of LLM into the CRL paradigm, facilitating  knowledge transfer across diverse tasks. 
    \item Extensive experiments in MiniGrid provide empirical evidence of \ourm{}'s effectiveness.
\end{itemize}

\section{Background}
\subsection{Preliminaries}
The continual reinforcement learning process can be formulated as a series of related Markov decision processes (MDPs) $\{<\mathcal{S}^i,\mathcal{A}^i, P^i, R^i>\}$.
Each MDP represents a different task or problem instance that an agent needs to solve over its lifetime.
Here, $\mathcal{S}^i$ and $\mathcal{A}^i$ denote the state and action space of task $i$, respectively, while $P^i: \mathcal{S}^i \times \mathcal{S}^i \times \mathcal{A}^i \to [0,1]$ is the transition probability function, and $R^i: \mathcal{S}^i \times \mathcal{A}^i \to [r^{\rm{min}},r^{\rm{max}}]$ is the reward function.
At each time step, the learning agent perceives the current state $s^i_t \in \mathcal{S}^i$ and selects an action $a^i_t \in \mathcal{A}^i$ according to its policy $\pi_\theta: \mathcal{S}^i \times \mathcal{A}^i \to [0,1]$ with parameters $\theta$.
The agent then transitions to the next state $s^i_{t+1} \sim P^i(s^i_t,a^i_t)$ and receives a reward $r^i_t = R^i(s^i_t,a^i_t,s^i_{t+1})$.
The target of an agent on the task $i$ is to maximize the expected return $ \mathbb{E} \left[ \sum_{t=0}^H \gamma^t  R^i(s^i_t, a^i_t, s^i_{t+1}) \right]$,
where $\gamma$ is the discount factor, and $H$ is the horizon.

\subsection{Related Works}
\subsubsection{Continual Reinforcement Learning}
CRL focuses on training RL agents to learn multiple tasks sequentially without prior knowledge, garnering significant interest due to its relevance to real-world AI applications \cite{khetarpal2022towards}.
A central issue in CRL is catastrophic forgetting, which has led to various strategies for knowledge retention.
PackNet and related pruning methods \cite{mallya2018packnet,schwarz2021powerpropagation} preserve model parameters but often require knowledge of task count.
Experience replay techniques such as CLEAR \cite{rolnick2019experience} use buffers to retain past experiences, but face memory scalability challenges.
In addition, some methods prevent forgetting by maintaining multiple policies or a subspace of policies \cite{schopf2022hypernetworkppocontinualreinforcementlearning,gaya2023buildingsubspacepoliciesscalablecontinual}.
Furthermore, task-agnostic CRL research indicates that rapid adaptation can also help prevent forgetting \cite{caccia2023taskagnosticcontinualreinforcementlearninggaining}.

Another issue in CRL is transfer learning, which is crucial for efficient policy adaptation.
Naive approaches, like fine-tuning, that train a single model on each new task provide good scalability and transferability but suffer from catastrophic forgetting.
Regularization-based methods, such as EWC \cite{kirkpatrick2017overcoming,wang2023comprehensive}, have been proposed to prevent this side effect, but often reduce plasticity.
Some architectural innovations have been proposed to balance the trade-off between plasticity and stability \cite{rusu2016progressiveneuralnetworks,berseth2022compscontinualmetapolicysearch}.
Furthermore, methods like OWL \cite{kessler2022samestatedifferenttaskcontinual} and MAXQINIT \cite{abel2018policyvaluetransferlifelongreinforcement} leverage policy factorization and value function transfer, respectively, for improved learning.

Most existing methods perform well when applied to sequences of tasks that exhibit high environmental similarity, such as tasks where only specific parameters within the environment are altered or the objectives within the same environment are different.
However, their effectiveness is greatly diminished when dealing with a sequence of diverse tasks.
Our proposed framework aims to overcome this limitation by leveraging multi-granularity knowledge transfer.

\subsubsection{Reinforcement Learning With the LLM}
Recent advancements have combined the LLM with reinforcement learning to address reinforcement learning challenges such as sample efficiency and generalization.
These integrations utilize LLM's vast knowledge and reasoning ability to improve agents' performance.
For instance, studies have developed methods for efficient agent-LLM interactions \cite{hu2023enablingintelligentinteractionsagentllm} and unified agent foundations with language models for better experience reuse \cite{palo2023unifiedagentfoundationmodels}.

Frameworks that merge LLMs with RL have also emerged, such as LAMP, which pretrains RL agents with Vision-Language Models \cite{adeniji2023languagerewardmodulationpretrainingreinforcement}, and HiP, which uses foundation models for long-horizon tasks \cite{ajay2023compositional}.
In addition, the RAFA framework offers a principled approach with guarantees to optimize reasoning and actions \cite{liu2023reasonfutureactnowprincipled}.
Research into LLM-guided skill learning and reward shaping is also gaining traction.
The BOSS method \cite{zhang2023bootstrapyourownskillslearning} and the work on LLM-based reward structures for robotics \cite{yu2023languagerewardsroboticskillsynthesis} exemplify this trend, with studies like \cite{kwon2023rewarddesignlanguagemodels} focusing on LLM-driven reward design.
Furthermore, VLA combines coarse-grained visual-language policies and fine-grained reinforcement learning policies, demonstrating significant zero-shot transfer performance in the real world \cite{yang2024learning}.

CRL diverges from the traditional RL paradigm by focusing on complex and dynamic environments where the agent constantly encounters a sequence of tasks. 
Unlike other research on LLM support for RL, which mainly deals with static tasks or single tasks, our work integrates LLM's powerful reasoning ability into the CRL paradigm for the first time. 
This distinction is crucial, as CRL poses unique challenges such as the necessity for high-level knowledge transfer across diverse tasks. 
Existing RL methods with LLM cannot address these challenges directly.
Our framework's novelty lies in its utilization of the LLM and the policy library not only to generate high-level policies that guide the learning process but also to facilitate high-level knowledge transfer between tasks, thus improving the agent's adaptability.

\section{Method}
In this section, we first clarify the research problem and the overall framework.
Then we elaborate on two layers of \ourm{}: (1) the coarse-grained policy formulation, and (2) the fine-grained policy learning.
Finally, we describe the policy library in \ourm{}.

\paragraph{Problem Formalization.}
Our goal is to use the LLM's powerful reasoning ability to enhance the capability of CRL agents.
We formalize the CRL problem in the goal-oriented RL framework \cite{chen2022near,colas2022autotelic}.
The agent perceives not only the current state of the environment but is also provided with a specific goal $g \in \mathcal{G}$, where $\mathcal{G}$ represents the space of possible goals.
For task $i$ and its corresponding goal sequence $\mathbf{g}^i=(g^i_1,\cdots,g^i_m)$, the agent's objective is to learn a policy $\pi^i(s^i)$ that effectively maps each state $s^i$ to an appropriate action $a^i$.
The aim of the agent is to maximize the expected cumulative reward over time, guided by the sum of the extrinsic reward $R^i(s^i, a^i, s^{\prime i})$ and the intrinsic rewards $(R^{\prime i}(s^i, a^i, s^{\prime i}, g^i_1), \cdots, R^{\prime i}(s^i, a^i, s^{\prime i}, g^i_m))$ \cite{nat2019intrinsic}, where $s^{\prime i}$ represents the next state in the task $i$.
The former is determined by the environment and is usually sparse.
The latter is typically defined so that the agent receives a reward if it achieves the given goal.

\begin{figure*}[ht]
    \begin{large}
        \begin{center}
            \includegraphics[width=1\textwidth]{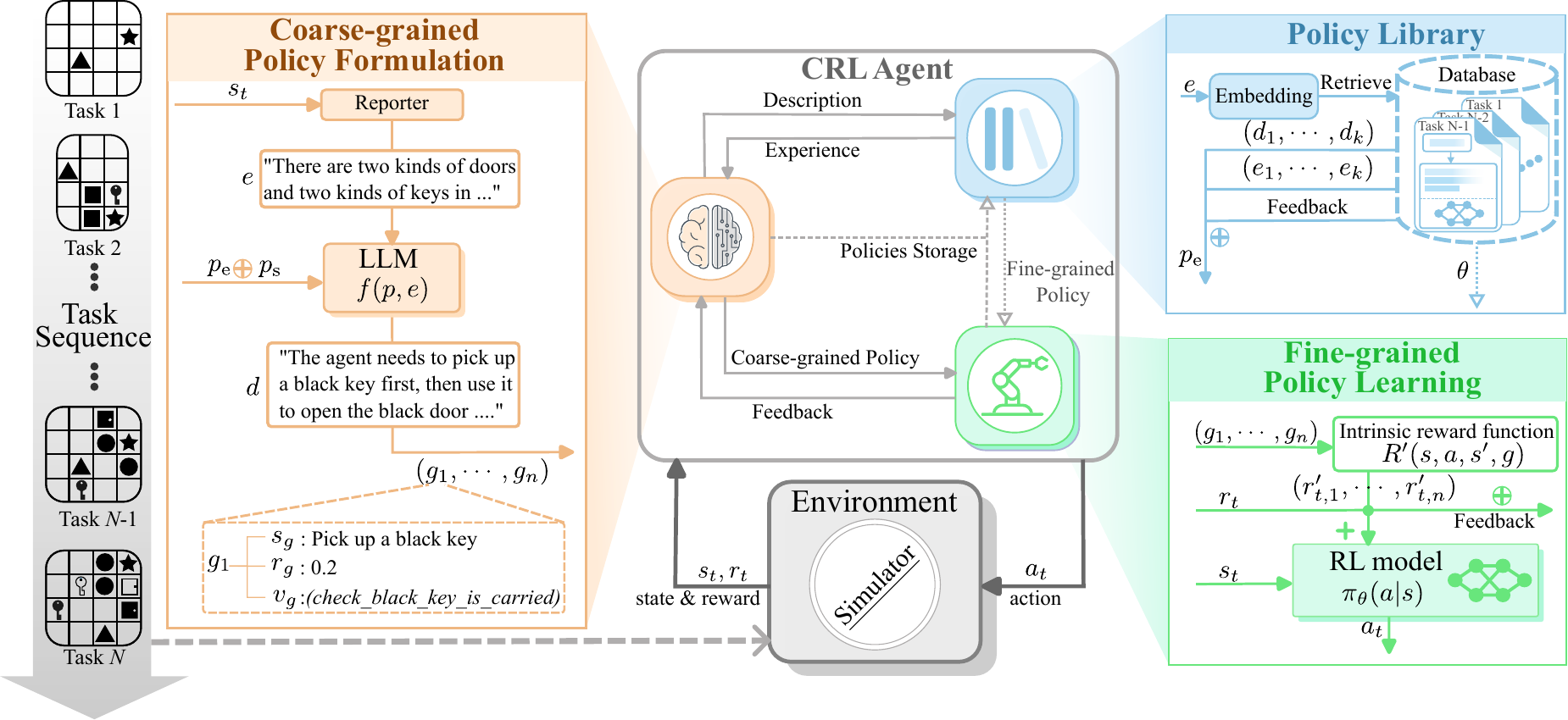}
        \end{center}
        \caption{The illustration of the proposed framework.
            The middle section depicts the internal interactions (\textbf{light gray line}) and external interactions (\textbf{dark gray line}) in \ourm{}.
            Internally, the agent is structured in two layers: the coarse-grained policy formulation (\textbf{orange}) and the fine-grained policy learning (\textbf{green}).
            Furthermore, the policy library (\textbf{blue}) is constructed to store and retrieve policies.
            The formulation of the coarse-grained policy is initiated when the agent interacts first with the environment.
            If the policy library is not empty, related experiences (including coarse-grained policies, corresponding environment descriptions and feedback) will be retrieved through the description of the environment to assist in the formulation.
            Once the coarse-grained policy is established, it guides the learning of the fine-grained policy.
            Through feedback from the RL learning process, the LLM gradually refines its coarse-grained policy to improve performance.
            Upon completion of a task, the successful policies are stored in the policy library.
        }
        \label{fig:framework}
    \end{large}
\end{figure*}

\paragraph{Overview.}
\ourm{} is illustrated in Figure \ref{fig:framework}.
For simplicity, we have omitted the task-indexed superscript in the description.
The framework orchestrates the interaction between the CRL agent and the environment simulator for each task in the task sequence, aligning with standard RL protocols.
For each task, the corresponding environment simulator provides the current state $s_t$ to the agent at each timestep, which then selects an action $a_t$ to perform, resulting in a reward $r_t$ and the next state.
Internally, the CRL agent is structured in two layers: the coarse-grained policy formulation and the fine-grained policy learning.
Additionally, a policy library is maintained for the storage and retrieval of policies.

\subsection{Coarse-Grained Policy Formulation}
To integrate LLM into the goal-oriented RL framework, we conceptualize LLM as a function $\mathcal{F}: \mathcal{P} \times \mathcal{E} \rightarrow \mathcal{D}$, where $\mathcal{P}$ is the set of all prompts, $\mathcal{E}$ is the set of textual descriptions of the environment, and $\mathcal{D}$ is the set of textual descriptions of the sequence of goals.
The description of the environment $e\in\mathcal{E}$ can be automatically obtained through a hard-coded reporter \cite{dasgupta2022collaboratinglanguagemodelsembodiedreasoning} or a vision-language model \cite{palo2023towards,yuqing2023guiding}.
The prompt $p \in \mathcal{P}$ is a concatenation of two parts: $p_{\rm{s}}$ and $p_{\rm{e}}$, where $p_{\rm{s}}$ is the system prompt and $p_{\rm{e}}$ is the experience retrieved from the policy library.
Description of each coarse-grained policy, that is, textual output $d \in \mathcal{D}$ is mapped to the corresponding goal sequence $\mathbf{g}\in \mathcal{G}^*$ through a defined parsing function.
In order to improve the stability of LLM's output, we structure the goal into a tuple: $g = (s_g, v_g, r_g)$, where $s_g \in \mathcal{E}$ is the textual description of the goal state, $v_g$ is a verification function $v_g: \mathcal{E} \times \mathcal{S} \rightarrow \{0, 1\}$, $r_g \in \mathbb{R}$ is a intrinsic reward value of the goal.
Optional validation functions are provided in $p_{\rm{s}}$.

The intrinsic reward can not only guide the agent to learn to achieve the corresponding goal but also serve as a metric for evaluating whether the agent has achieved the goal.
This metric is continuously monitored during the interactions of the CRL agent with the environment.
When this average exceeds a predefined threshold, it is taken as an indication that the agent has successfully achieved the goals.
Consequently, the LLM shifts its focus to the next goal in the sequence, and the process repeats.
The iterative process of coarse-grained policy formulation, validation, and improvement draws inspiration from the concept of Reflexion \cite{shinn2023reflexion}.
In instances where the agent fails to meet the threshold, indicating that the goal has not been met, the LLM is prompted to adjust the coarse-grained policy.
This adjustment is informed by feedback from the agent's performance.
The LLM is first required to analyze the feedback to pinpoint the reasons behind the shortfall in the previous policy.
With powerful reasoning ability, the LLM then formulates a new coarse-grained policy aimed at overcoming the identified challenges.

\subsection{Fine-Grained Policy Learning}
Fine-grained policy learning is guided by the formulated coarse-grained policy.
The RL model learns the fine-grained policy using raw state input $s_t$ and total intrinsic reward $r^{\prime}_t$ based on the goal sequence $\mathbf{g}=(g_1, \cdots,g_m)$.
The agent's intrinsic reward function is defined as:
\begin{equation}
    R^{\prime}(s, a, s', g) = \begin{cases} r_g, & \text{if } v_g(s_g,s') = 1  \\ 0, & \text{otherwise} \end{cases}.
    \label{eq: intrinsic}
\end{equation}
This reward structure provides a more granular view into the effectiveness of goal-related actions, encouraging the agent to learn a policy that is aligned with the goal more efficiently.
Furthermore, the state-dependent design of this function improves the generalizability of coarse-grained policies. Then, the internal rewards for each goal $\{r^{\prime}_{t,l}=R^{\prime}(s_i, a_i, s_{i+1}, g_l) \vert l = 1,\cdots, m\}$ and the external reward $r_t$ from the environment will be textual and will be concatenated as the feedback.
The total intrinsic reward is $r^{\prime}_t = \sum_{l=1}^{m} r^{\prime}_{t,l}$.
Therefore, the aim of RL agent with goal sequence $g$ can be expressed as
\begin{equation}
    \begin{aligned}
        max_{\theta}  J(\theta,g)   =                                                          \mathbb{E}_{\pi_{\theta}} \left[ \sum_{t=0}^H \gamma^t ( r_t+r^{\prime}_t) \right],
    \end{aligned}
    \label{eq: rl_goal}
\end{equation}
where $\theta$ is the parameter of fine-grained policy.

\subsection{Policy Library}
Recent studies \cite{wang2023voyageropenendedembodiedagentlarge,zhang2023bootstrapyourownskillslearning} have highlighted the potential of LLMs' reasoning and planning ability in facilitating continual learning (CL) for agents, which involves continuous acquisition and update of skills \cite{wang2023comprehensive}.
Two core challenges in continual learning are catastrophic forgetting and knowledge transfer \cite{hadsell2020embracing,mccloskey1989catastrophic}.
One method to address the these challenges is to design a knowledge base \cite{li2023cl} or a skill library to store knowledge or skills\cite{wang2023voyageropenendedembodiedagentlarge}.
By using validated successful experiences, agents can not only reduce catastrophic forgetting but also rapidly enhance their capabilities.

Based on the above findings, \ourm{} maintains a policy library to store the policies that solve tasks successfully.
The description of each coarse-grained policy $d$ and the corresponding feedback is indexed by the embedding of the environment description $e$ of its corresponding task, which can be retrieved in similar tasks in the future.
The policy library can be continuously extended throughout the life of the CRL agent.
Given a new environment description $e$, \ourm{} can use embedding to retrieve the first $k$ similar environment descriptions and concatenate them with goal descriptions and corresponding feedback as experience $p_{\rm{e}}$.
Specific policies for new tasks can be learned under the guidance of coarse-grained goals.
Specifically, We adopt a vector similarity search approach, with cosine similarity as the metric. 
When the number of tasks is small, simple storage strategies can also be used for the policy library.
In addition, for each environment description $e$, \ourm{} also stores the parameter of the fine-grained policy $\theta$ that has been learned.
This can accelerate learning on similar tasks and further reduce catastrophic forgetting if storage space is sufficient.
Even if the storage space is strict, only storing coarse-grained policies (text) can perform knowledge transfer with small memory usage.
By storing both coarse-grained and fine-grained learning policies in the policy library, the agent's capabilities for diverse tasks can be enhanced over time.

\section{Experiments}
In this section, we evaluate our framework in several continual reinforcement learning tasks. 
More details about the experimental implementation are provided in the supplementary material.

\subsection{Environment}
For our experiments, we utilized a suite of MiniGrid environments \cite{MinigridMiniworld23} to evaluate the efficacy of \ourm{} in addressing CRL tasks.
These environments feature image-based state observations, a discrete set of possible actions, and various objects characterized by their color and type. 
We intentionally simplified the state and action spaces to streamline the policy learning process, enabling more rapid experimental iterations. 
This approach allowed us to focus on the core objective of this study: leveraging hierarchical knowledge transfer for CRL. 
We crafted a sequence of four distinct tasks within the MiniGrid framework. 
This sequence is arranged in ascending order of difficulty and aligns with the human learning process, facilitating better tracking of knowledge transfer throughout the learning process. 
A comprehensive account of these modified environments, including their specific configurations and visual representations, is available in supplementary material.

\subsection{Experiment Setup}\label{sec: setup}

\paragraph{Baselines.} We consider the following baselines: 
\begin{itemize}
    \item \textbf{Single-Task (SG)}: This baseline represents a traditional RL setup where a distinct agent is trained exclusively on each task.
          There is no sharing or transfer of knowledge between tasks in this method.
          This serves as a foundational comparison point to underscore the advantages of CL, as it lacks any mechanism for knowledge retention or transfer.
    \item \textbf{Fine-Tuning (FT)}: Building upon the standard RL algorithm, this baseline differs from SG by using a single agent that is sequentially fine-tuned across different tasks.
            The agent trained on the preceding task is adapted for the subsequent task, effectively using the prior model as a starting point.
          This baseline provides a basic measure of an agent's capacity to maintain knowledge of earlier tasks while encountering new tasks.
    \item \textbf{Fine-Tuning with L2 Regularization (FT-L2)}:
          Based on FT, this baseline incorporates L2 regularization during the fine-tuning process.
            The addition of L2 regularization aims to mitigate catastrophic forgetting by penalizing significant changes to the weights that are important for previous tasks.
            This helps in preserving performance on earlier tasks while learning new ones.
          FT-L2 can be viewed as the simplest implementation of regularization-based CRL methods \cite{kirkpatrick2017overcoming,wang2023comprehensive}.
    \item \textbf{PackNet} \cite{mallya2018packnet}:
          This baseline is a representative CRL method based on parameter isolation \cite{wang2023comprehensive}.
          It utilizes the network pruning technique to efficiently allocate the neural network's capacity throughout the continual learning of tasks.
            After training on a task, PackNet identifies the crucial weights to retain, prunes the less essential ones, and thus creates room for the network to learn additional tasks.
            PackNet is a representative CRL method based on parameter isolation \cite{wang2023comprehensive}.
\end{itemize}
Note that our method is not directly comparable with certain methods, such as CLEAR \cite{rolnick2019experience} and ClonEx-SAC \cite{wolczyk2022disentangling}, as it requires access to the data from previous tasks for training.
Our framework can continually learn without the need to revisit past task data, which is a common constraint in real-world applications.

\paragraph{Metrics.} To evaluate the effectiveness of \ourm{}, we use the normalized average return to measure the performance of the trained agents.
Following standard practice in supervised continual learning literature \cite{wolczyk2021continual,li2024learning}, we define a suite of metrics based on the agent's performance throughout different phases of its training process.
Based on the agent's normalized average return, we evaluated the continual learning performance of our framework and baselines using the following metrics: \textit{average performance}, \textit{forward transfer} and \textit{forgetting}, as detailed in the supplemental material.

\subsection{Competitive Results}
To evaluate our framework, we compare \ourm{} with other baselines across four challenging tasks that have the same state and action space.
Each experiment is trained in 5M steps and replicated with five random seeds of environments to ensure reliability.

\begin{figure}[tp]
    \centering
    \includegraphics[width=0.48\textwidth]{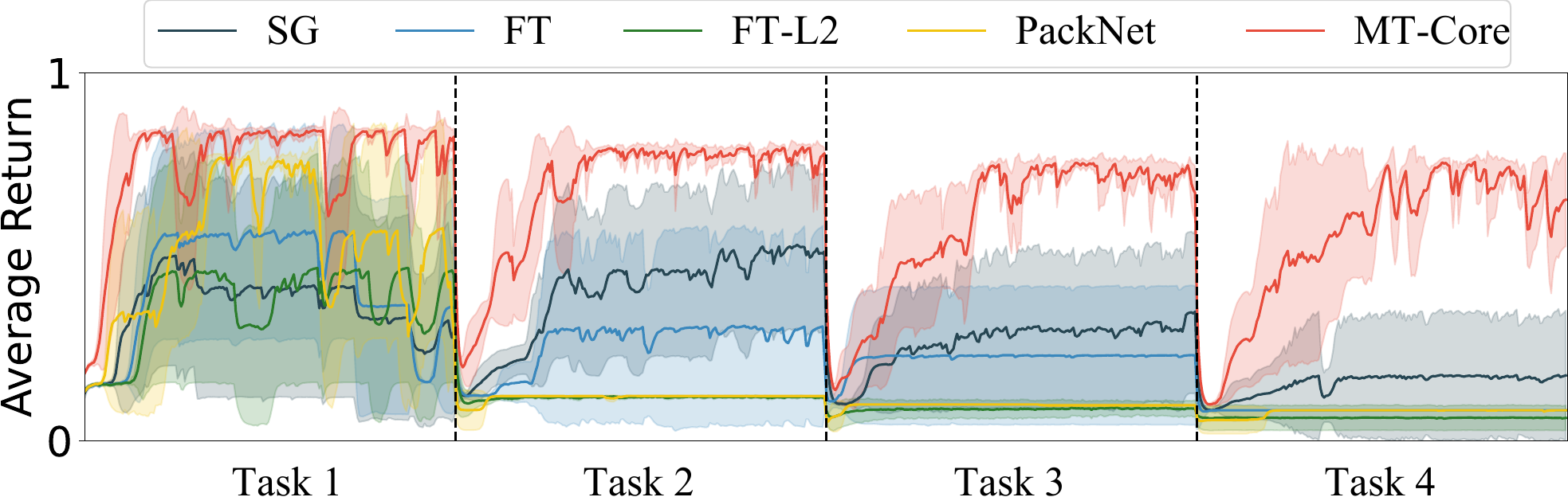}
    \caption{Performance during training across four tasks (smoothed
with exponential moving average). The x-axis represents the training progress for each task. The y-axis represents the normalized
average return of each task. The shaded area depicts the means $\pm$
the standard deviation over five random seeds.}
    \label{fig: performance}
\end{figure}

\noindent\textbf{Overall Performance.} As illustrated in Figure \ref{fig: performance}, the baselines perform well on the first Task.
However, their performance gets worse progressively as the difficulty of the tasks increases.
A common trend of performance degradation in the transition between tasks is evident across all methods.
Specifically, the SG baseline, while effective on the first task, is unable to maintain its performance due to its inherent limitation.
Other baselines, such as FT, FT-L2, and PackNet, also start strong on the first task but encounter difficulties in the subsequent tasks.
This phenomenon suggests that a singular shared policy may not be robust enough to handle a sequence of highly varied tasks.
The balance between policy stability and plasticity is critical, with higher stability often leading to reduced adaptability, as observed in the performance degradation of these baselines.
In contrast, \ourm{} consistently surpasses the other baselines, showcasing remarkable performance across all tasks.
Although it is not immune to performance degradation when transitioning between tasks with significant differences, it demonstrates a rapid recovery in performance.
This can be attributed to its multi-granularity structure, which leverages the powerful capabilities of LLM and promotes the transfer of coarse-grained policy knowledge, thereby enhancing the agent's ability to adapt to new and diverse tasks.

\begin{table}
    \begin{small}
        \centering  
        \begin{tabular}[c]{l|ccccc}
            \hline
            Metric           & SG     & FT      & FT-L2           & PackNet & \ourm{}         \\
            \hline
            $A_N(\uparrow)$  & $0.30$ & $0.25$  & $0.16$          & $0.20$  & $\textbf{0.65}$ \\
            $FW(\uparrow)$   & $0.00$ & $-0.05$ & $-0.14$         & $-0.09$ & $\textbf{0.17}$ \\
            $FG(\downarrow)$ & $0.12$ & $0.09$  & $\textbf{0.05}$ & $0.06$  & $0.12$          \\
            \hline
        \end{tabular}
        \caption{CL performance of \ourm{} and four baselines.}
        \label{tab: cl_metric}
    \end{small}
\end{table}

\noindent\textbf{Continual Learning Performance.}
Table \ref{tab: cl_metric} shows the evaluation results in terms of CL metrics.
The average performance metric clearly demonstrates \ourm{}'s superiority in task sequence with diverse tasks, significantly outperforming other baseline methods.
The forward transfer metric is particularly important, as it measures the capacity of an agent to utilize knowledge from previous tasks.
All baselines, with the exception of \ourm{}, present forward transfer metrics of less than or equal to zero.
In addition, the average performace of \ourm{} is more than twice that of the best baseline.
This suggests that traditional approaches struggle to transfer previously acquired knowledge to new tasks effectively.
Baselines that rely on regularization strategies, such as FT-L2, or parameter isolation techniques, like PackNet, may even hinder the learning of new tasks.

In contrast, \ourm{} exhibits positive transfer performance, underscoring the benefits of transferring coarse-grained policies.
When examining the forgetting metrics, FT, FT-L2, and PackNet show relatively low scores.
This can be attributed, in part, to their emphasis on model stability.
However, a more critical factor is their inherently lower performance on subsequent tasks, such as Task 2, Task 3, and Task 4.
Although \ourm{} does not achieve the lower score in the forgetting metric, its exceptional forward transfer capability and strong average performance accentuate its proficiency in handling diverse task sequences.
Note that \ourm{} does not resort to fine-tuning strategy in experiments to focus on coarse-grained knowledge transfer, although it could be employed in situations where minimizing forgetting is a priority.

\subsection{Heterogeneous Tasks Experiments}

\begin{figure}[tp]
    \centering 
    \subfloat[]{
        \includegraphics[width=0.205\textwidth]{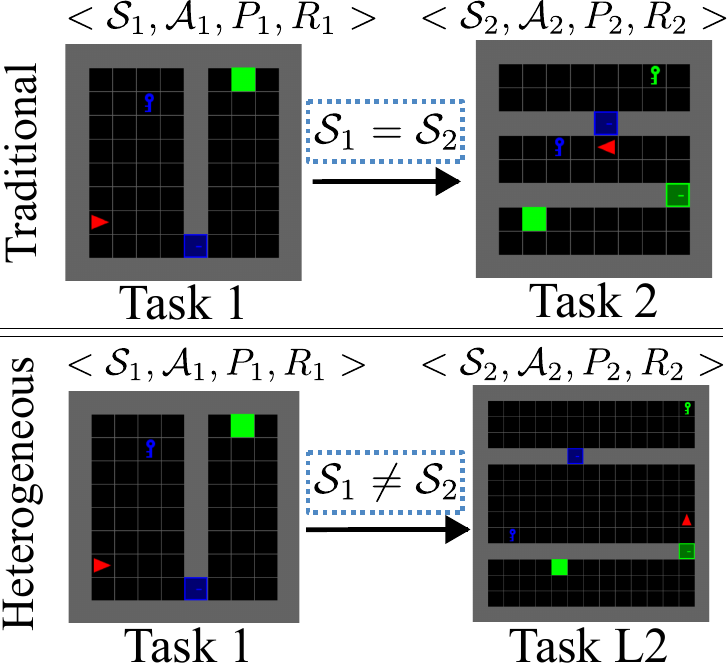}
        \label{fig: hcrl1}
    }
    \hfill
    \subfloat[]{
        \includegraphics[width=0.235\textwidth]{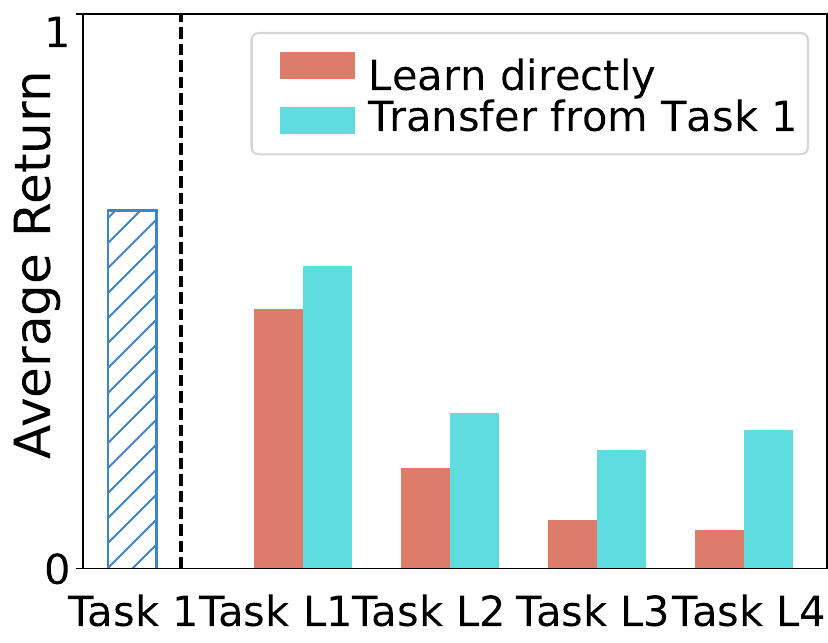}
        \label{fig: hcrl2}
    }
    \hfill
    \caption{Heterogeneous tasks experiments. \textbf{(a)} A simple illustration. The setting of traditional CRL requires the same state spaces of the tasks (\textbf{above}). In our heterogeneous task experiments, the state space of each task may be different (\textbf{below}). \textbf{(b)} Experiment results. The left side of the dashed line represents the performance on the original Task 1. The right side of the dashed line represents the performances on tasks with larger state space.
    }
    \label{fig: herl}
\end{figure}

Most existing works have focused on the transfer ability of CRL agents in environments that have the same state space.
In contrast, our study explores the application of CRL agents to heterogeneous tasks, where environments have different state spaces or action spaces.
For the simplification of the problem, we concentrate on tasks with differing state spaces.

As depicted in Figure \ref{fig: hcrl1}, heterogeneous tasks present a unique challenge.
While it is possible to accommodate differences in the size of the state space during knowledge transfer by employing a masking technique, this approach necessitates prior knowledge of the maximum state space size, which is impractical for RL agents expected to learn continually in dynamic environments.
On the contrary, the coarse-grained policy transfer of \ourm{} can be applied to heterogeneous tasks without the need for masking.

In our experiments, we evaluate the effectiveness of \ourm{} across two heterogeneous tasks.
The first task has a relatively small state space, as shown in the left below of Figure \ref{fig: hcrl1}.
In contrast, the second task has a larger state space, such as the right below of the same figure.
Figure \ref{fig: hcrl2} reports the results of these experiments.
The first task is Task 1, while the second task is Task L1, Task L2, Task L3, or Task L4, each representing a larger state space variant of Task 1 through Task 4.
Direct learning agents who are exposed to tasks with large state spaces from the outset may struggle to learn effective policies within the set number of steps, resulting in poor performance.
This phenomenon is more obvious on more difficult tasks, such as Task L3 and Task L4.
However, agents that are learned on Task 1 can transfer coarse-grained knowledge to subsequent tasks.
Despite the increasing difficulty of these tasks, the agents still demonstrate comparable performance.
This evidence suggests that \ourm{} retains transfer ability even when applied to heterogeneous tasks.

\subsection{Ablation Study}

\begin{figure}[tp]
    \begin{small}
        \begin{center}
            \subfloat[]{\includegraphics[width=0.47\textwidth]{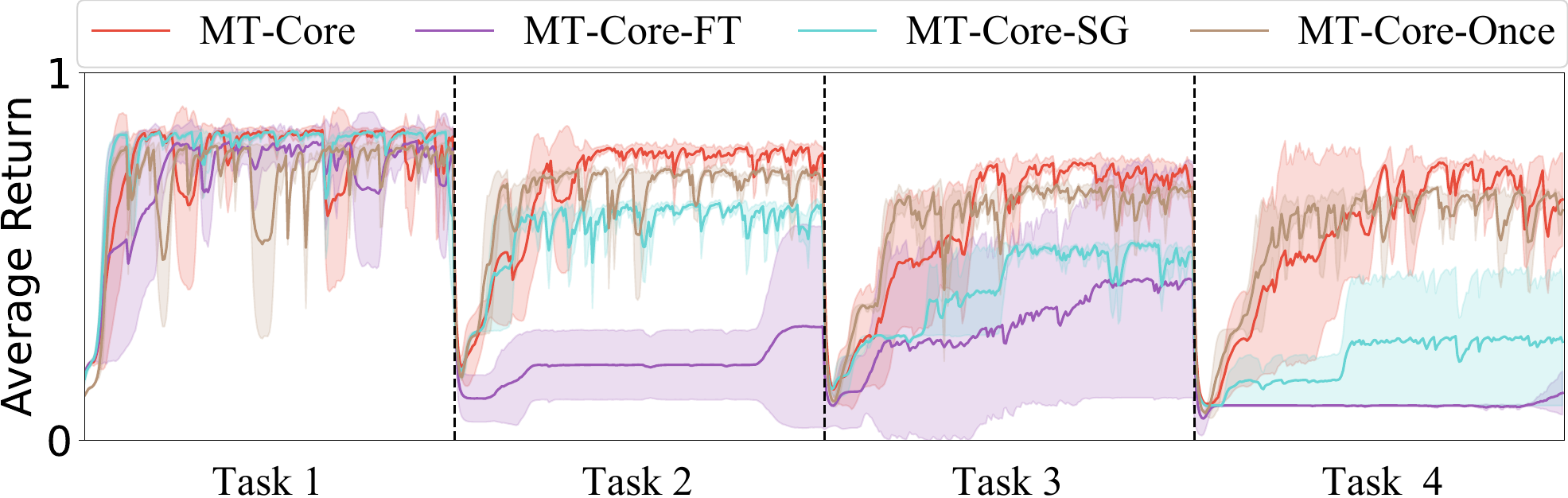} \label{fig: ablation}}
            \hfill
            \subfloat[]{\includegraphics[width=0.47\textwidth]{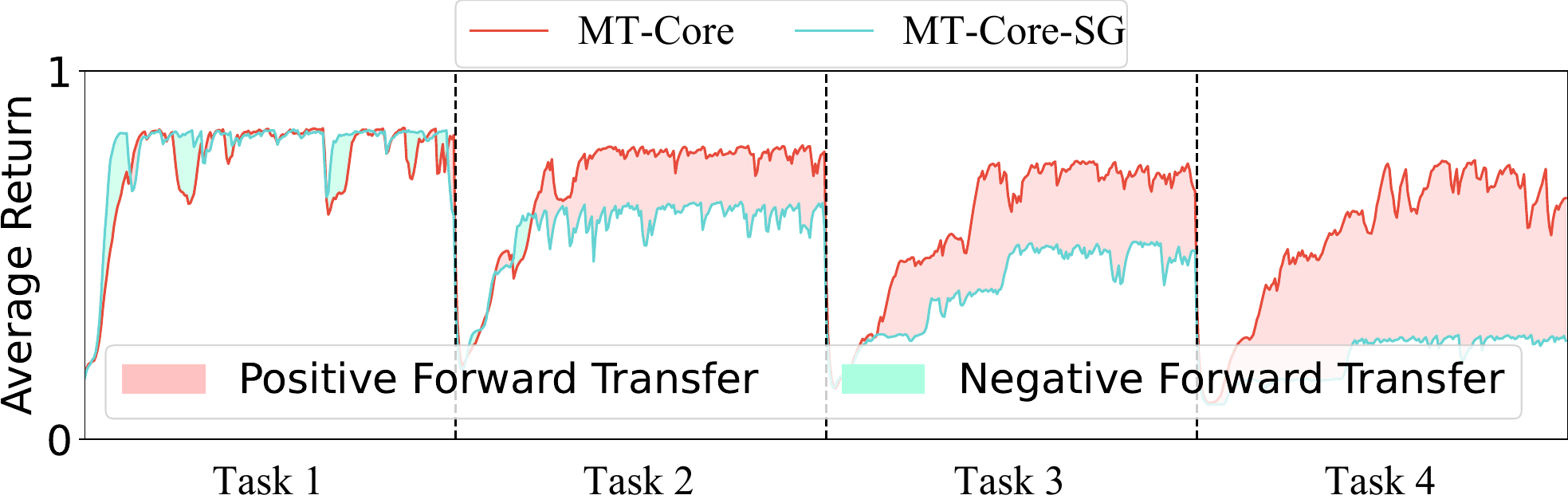} \label{fig: fwt}}
        \end{center}
        \caption{Ablation study results. \textbf{(a)} Performance of \ourm{} and its variants during training.
        \textbf{(b)} Forward transfer performance of
        \ourm{} during training. The red areas between the curves represent positive forward transfer and green represent negative forward transfer.
        }
        \label{fig: ablation_performance}
    \end{small}
\end{figure}

\begin{table}[ht]
    \begin{small}
        \centering 
        \begin{tabular}[c]{l|ccc}
            \hline
                             & $A_N(\uparrow)$ & $FW(\uparrow)$  & $FG(\downarrow)$ \\
            \hline
            SG           & $0.30$          & $0.00$             & $0.10$           \\
            \ourm{}      & $\textbf{0.65}$ & $\textbf{0.17}$ & $0.12$           \\
            \ourm{}-FT   & $0.33$          & $-0.14$         & $\textbf{0.09}$  \\
            \ourm{}-SG   & $0.47$          & $0.00$             & $0.19$           \\
            \ourm{}-Once & $0.62$          & $0.15$          & $0.13$           \\
            \hline
        \end{tabular}
        \caption{CL performance of the variants of \ourm{}.}
        \label{tab:ablation_results}
    \end{small}
\end{table}

In this part, we conduct an ablation study to investigate what affects \ourm{}'s performance in CRL.
We consider several variants of \ourm{} to understand the effects:
\begin{itemize}
    \item \textbf{\ourm{}-FT}: \ourm{} fine-tuning the policy from the previous task when encountering a new task.
    \item \textbf{\ourm{}-SG}: \ourm{} without policy library.
          This variant treats each task as an isolated learning problem, akin to training single agents for each task.
    \item \textbf{\ourm{}-Once}: \ourm{} without feedback mechanisms.
          In this variant, the coarse-grained policy formulation of LLM is performed only once for each task.
\end{itemize}

The results of our study are presented in Table \ref{tab:ablation_results} and Figure \ref{fig: ablation_performance}.
For better comparison, we also include the performance of the SG baseline in the Table.
Our findings are as follows:
Firstly, \ourm{}-SG shows superior performance compared to the SG baseline, which confirms that the LLM's powerful capabilities can help RL.
An RL agent can take advantage of reasoning ability in the LLM to significantly reduce the difficulties in solving these tasks.
Secondly, while \ourm{}-SG outperforms the baseline, it falls short of the average performance and forward transfer metrics achieved by the \ourm{}.
The gap between \ourm{} with \ourm{}-SG emphasizes the value of the policy library in facilitating knowledge transfer, ultimately boosting the agent's overall performance (see the figure in \ref{fig: fwt}).
Furthermore, \ourm{}-Once exhibits slightly lower performance on Task 2, Task 3 and Task 4 than \ourm{}.
This phenomenon suggests that feedback can improve the performance of LLM in dealing with these challenging tasks.
Lastly, the \ourm{}-FT variant underperforms on all tasks except Task 1.
This observation aligns with the previous finding in competitive results that methods designed to enhance policy stability may decrease the agent's performance on subsequent tasks when those tasks differ significantly from one another.

\begin{figure}[tp]
    \begin{small}
        \centering
        \subfloat[]{\includegraphics[width=0.11\textwidth]{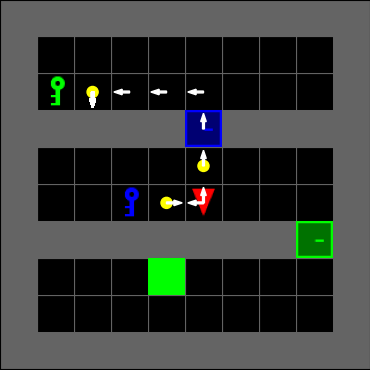} \label{fig: example}}
        \hfill
        \subfloat[]{\includegraphics[width=0.11\textwidth]
            {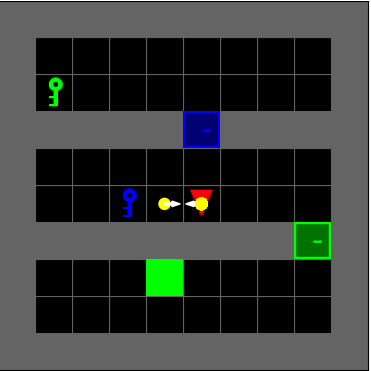} \label{fig: sg_example}}
        \hfill
        \subfloat[]{\includegraphics[width=0.11\textwidth]
            {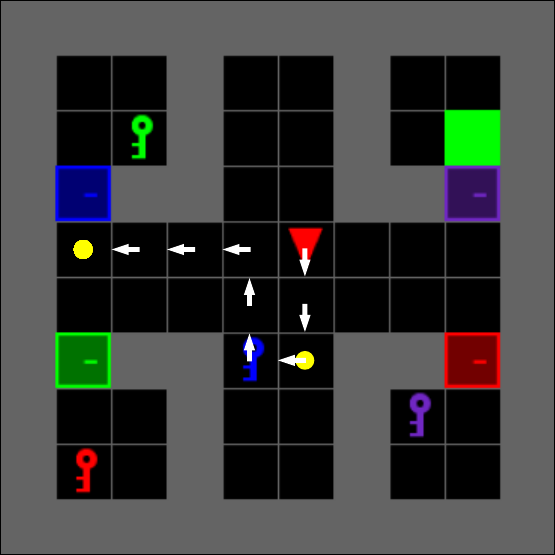} \label{fig: example4}}
        \hfill
        \subfloat[]{\includegraphics[width=0.11\textwidth]
            {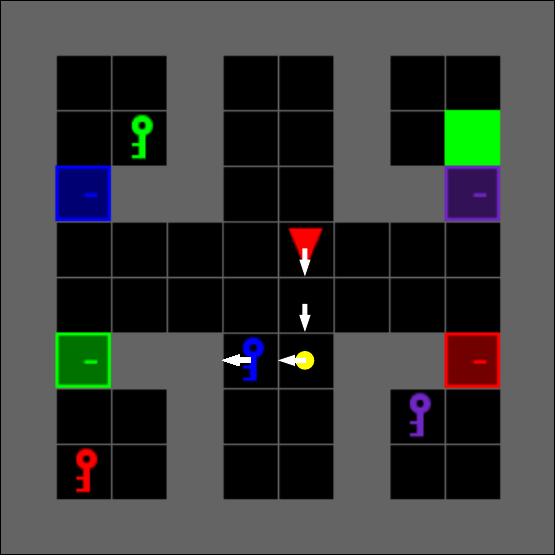} \label{fig: sg_example4}}
        \caption{
            Early-stage policy visualization for agents on Task 2 (0.5M steps) and Task 4 (1M steps), following learning on Task 1 (5M steps).
            \textbf{(a) and (c)} depict the policies learned by \ourm{} for Task 2 and Task 4 respectively, while \textbf{(b) and (d)} represent the policies by \ourm{}-SG for the same tasks.
            The forward actions of agents are denoted by white arrows, while the yellow dots denote interactive actions such as pickup and toggle.
            For the clearness, steering actions are omitted from these visual representations.
        }
        \label{fig: sl_examples}
    \end{small}
\end{figure}

\noindent\textbf{Policy Visualization.} Figure \ref{fig: sl_examples} further illustrates the difference between \ourm{} and \ourm{}-SG on Task 2 and Task 4.
As shown in Figure \ref{fig: example}, the agent of \ourm{} demonstrates the capability to swiftly learn to pick up the blue key and unlock the corresponding door.
In contrast, the agent of \ourm{}-SG struggles to progress beyond the initial room (Figure \ref{fig: sg_example}).
This performance gap can be attributed to the previous learning of the \ourm{} agent on Task 1, which shares common elements with Task 2, such as the blue key and door.
By drawing on previous experience, \ourm{} formulate relevant initial goals, accelerating learning on Task 2.
Similarly, when comparing Figure \ref{fig: example4} and Figure \ref{fig: sg_example4} for Task 4, we observe a consistent pattern.
Although Task 4 presents a higher level of difficulty, the agent with pre-acquired coarse-grained knowledge exhibits more effective behaviors than the agent learning from scratch.
\section{Conclusion}
In this work, we took a step in the direction of improving the generalization ability of CRL agents.
We investigated the limitations of existing CRL methods in transferring fine-grained knowledge across diverse tasks and proposed \ourm{}, a novel framework that aligns with the multi-granularity structure of human cognitive control, enhancing the coarse-grained knowledge transfer of CRL.
\ourm{} leverages a LLM for coarse-grained policy formulation, employs goal-oriented RL for fine-grained policy learning, and constructs the policy library to store and retrieve policies.
It addresses the challenge of learning across a sequence of diverse tasks by transferring coarse-grained knowledge, which has been demonstrated in our experiments.
Further results indicate that \ourm{} can be applied to heterogeneous tasks and achieves comparable transfer performance.

\section*{Acknowledgements}
This work was supported by the National Natural Science Foundation of China (Nos. 62476228), the Sichuan Science and Technology Program (Nos. 2024ZYD0180), and the Graduate Representative Achievement Cultivation Project of Southwest University of Finance and Economics (Nos. JGS2024068).

\bibliographystyle{named}
\bibliography{main}



\end{document}